\documentclass[a4paper,11pt]{article}

\usepackage{caption}
\usepackage{subcaption}

\usepackage[T1]{fontenc}
\usepackage[utf8]{inputenc}
\usepackage{graphicx}
\usepackage{xcolor}
\usepackage{float}
\usepackage{hyperref}
\usepackage{amsmath,amssymb,amsthm,textcomp}
\usepackage{enumerate}
\usepackage{multicol}
\usepackage{tikz}
\usepackage{geometry}
\usepackage[linesnumbered,ruled]{algorithm2e}
\usepackage{hyperref}
\begin{document}

\title{Adaptive PCA for Time-Varying Data}

%

\author{
  Salaheddin Alakkari and John Dingliana\\
  Graphics Vision and Visualisation Group (GV2)\\
  School of Computer Science and Statistics\\
  Trinity College Dublin\\
 \href{mailto:alakkars@tcd.ie}{alakkars@tcd.ie}|\href{mailto:John.Dingliana@tcd.ie}{John.Dingliana@tcd.ie}
}  
 \date{}
 
\maketitle

\begin{abstract}

    In this paper, we present an online adaptive PCA algorithm that is able to compute the full dimensional eigenspace per new time-step of sequential data. The algorithm is based on a one-step update rule that considers all second order correlations between previous samples and the new time-step. Our algorithm has $\mathcal{O}\left(n\right)$ complexity per new time-step in its deterministic mode and $\mathcal{O}\left(1\right)$ complexity per new time-step in its stochastic mode. We test our algorithm on a number of time-varying datasets of different physical phenomena. Explained variance curves indicate that our technique provides an excellent approximation to the original eigenspace computed using standard PCA in batch mode. In addition, our experiments show that the stochastic mode, despite its much lower computational complexity, converges to the same eigenspace computed using the deterministic mode.
    
\end{abstract}

\section{Introduction}

Principal Component Analysis (PCA) is one of the most important machine learning techniques for many reasons. Firstly, it is the only unsupervised learning algorithm that is theoretically proven to capture the maximal variability (information) of the input data given a fixed-size low-dimensional space. Another main reason is that it directly deals with the eigenspace of the problem on hand. In the real world, an endless amount of problems and physical phenomena can be modelled by eigenvalue equations. One important example is the Dirac equation which assumes that all variables of a physical object (speed, acceleration, etc) obey an eigenvalue problem \cite{rovelli2007quantum}. In quantum physics, the quantum states that an electron in an atom can take (labeled as 1S, 2S, 2P etc) are actually time-dependent eigenfunctions which are called ``quantum eigenstates'' \cite{smith2010waves}.

Despite the elegance of PCA, it has not been widely used until the last four decades. One reason for this is that, in its basic form, it has a quadratic space and time complexity which requires large memory and processing speed. Nowadays machines are shown to be more capable of handling such complexity thanks to larger available memory and faster CPU and GPU (Graphics Processing Unit) capabilities. However, for a wide range of problems where the dimensionality of the data is massive (due to the size and number of samples), extracting the principal components in the standard way becomes infeasible. Many algorithms have been developed to find the most significant principal components with linear complexity dependence on data size. However most of these approaches are stochastic and are limited to extracting a certain number of eigenvectors (principal components).

In this paper, we consider time-dependent systems that require regular monitoring and analysis for each new time-step. This is particularly important, for instance, in equilibria and stability analysis of the system. In many physical phenomena, such as the electron eigenstates example mentioned above, the time-dependent behavior of the significant eigenvectors converges to an equilibrium eigenstate. We propose an adaptive PCA algorithm that is able to capture all eigenvectors of the data and has $\mathcal{O}\left(n\right)$ complexity per new time-step in its deterministic mode and $\mathcal{O}\left(1\right)$ complexity per new time-step in its stochastic mode, where $n$ is the number of previous time-steps. We test this algorithm on six time-varying datasets of different physical phenomena. We compare the performance of our algorithm with the standard PCA applied in batch-mode.

\section{Background and Related Work}
	
	In the literature, there are two main directions that PCA research has taken. The first is that concerning applications which employ PCA for solving real-world problems and the second is that in the direction of PCA-optimization which is concerned with the optimization of the computational complexity of PCA. The link between the two directions is not clear since most studies in the application direction assume a pre-computed eigenspace and focus mainly on the distribution of test data in that eigenspace. On the other hand, in the optimization direction, the target use-case is not obvious. In addition, most of the optimization-direction algorithms are of a stochastic nature and are usually tested on rather simple datasets or data where a global eigenspase can be easily derived. In such a case, one can always consider a pre-computed eigenspace no matter what computational complexity was required for finding it. In fact, many online datasets provide a list of the most significant eigenvectors of the studied samples.
	
	With regard to the applications research, the use of PCA has been well reported in the fields such as Computer Vision and Computer Graphics. For instance, in facial recognition, Kirby and Sirovich \cite{kirby1990application}
	proposed PCA as a holistic representation
	of the human face in 2D images by extracting few orthogonal dimensions
	which form the face-space and were called eigenfaces \cite{turk1991eigenfaces}. Gong et al. \cite{gong1996investigation} were the first to find the relationship between the distribution of samples in the eigenspace, which were called manifolds, and the actual pose in an image of a human face. 
	The use of PCA was extended using Reproducing Kernel Hilbert Spaces which non-linearly map
	the face-space to a much higher dimensional space (Hilbert space)
	~\cite{yang2002kernel}.  Knittel and Paris~\cite{knittel09pcaseeding} employed a PCA-based technique to find initial seeds for vector quantization in image compression.  There are a number of previous reported uses of PCA-related methods in the computer graphics and visualization literature. For instance, Nishino et al. \cite{nishino1999eigen} proposed a method, called \emph{Eigen-texture}, which creates a 3D image from a sample of range images using PCA. They found that partitioning samples into smaller cell-images improved the rendering of surface-based 3D data.
	Grabner et al. \cite{grabner2003} proposed a hardware accelerated technique that uses the multiple eigenspaces method~\cite{leonardis2000} for image-based reconstruction of a  3D polygon mesh model. Liu et al.~\cite{liu2014} employed PCA for dynamic projections in the visualization of multivariate data. Broersen et al.~\cite{Broersen2005} discussed the use of PCA techniques in the generation of transfer functions, which are used to assign optical properties such as color and opacity to attributes in volume data. Takemoto et al.~\cite{Takemoto2013} used PCA for feature space reduction to support transfer funtion design and exploration of volumetric microscopy data. Fout and Ma~\cite{fout07} presented a volume compression method based on transform coding using the Karhunen-Lo\`eve Transform (KLT), which is closely related to PCA.
		
	In the PCA-optimization research, the power iteration remains one of the most popular techniques for finding the top $p$ eigenvectors \cite{golub2012matrix}. In the recent leterature, Shamir proposed a stochastic PCA algorithm that is proven to converge faster than the power iteration method \cite{shamir2015stochastic}. Both techniques have a lower bound complexity of $\mathcal{O}\left(n\log\left(\frac{1}{\epsilon}\right)\right)$ where $\epsilon$ is the precision of convergence. In addition, both techniques were experimentally tested to extract only a limited number of significant eigenvectors. Arora and De Sa et al.~\cite{ arora2012stochastic, arora2013stochastic,de2014global} proposed stochastic techniques that are based on the gradient-descent learning rule. The slow convergence rate of the gradient-descent rule is one main limitation of these techniques. Many algorithms were developed to find eigenvectors incrementally per new number of time-steps. Such techniques are referred to as incremental PCA algorithms. The update schemes proposed by Krasulina \cite{krasulina1969method} and Oja \cite{oja1982simplified,oje1983subspace} are the most popular incremental PCA techniques. Given a new time-step $x_{n+1}$ and a significant eigenvector $v$ for previous samples, the general update rule according to Oja's method~is \[
v^{i+1}=v^{i}+\alpha \left\langle x_{n+1},v^{i}\right\rangle x_{n+1};\;v^{i+1}=\frac{v^{i+1}}{\left\Vert v^{i+1}\right\Vert },
\] where $\alpha$ is the learning rate. This process will keep updating until converging to a stable state. The speed of convergence of this technique is a matter of ongoing research. Balsubramani et al. \cite{balsubramani2013fast} found that speed of convergence depends on the learning rate $\alpha$. Another problem with this technique (as we will find later in this study) is that it does not consider change in weightings of previous time-steps. Mitiagkas et al. proposed an incremental PCA algorithm for streaming data with computational complexity of $\mathcal{O}\left(n\log\left(n\right)\right)$~\cite{mitliagkas2013memory}.
	
	One important point to highlight is that most studies in both directions focus mainly on the most significant eigenvectors with little attention paid to the least significant ones. In fact, finding such eigenvectors was shown to play a key role in detecting outliers and non-belonging samples since they are perpendicular to the best fitting hyperplane. Jollife \cite{jolliffe2002principal} pointed out in his book that the principal components corresponding to the smallest
eigenvalues (variances) are not ``unstructured left -overs'' after
extracting the higher PCs and that they can be useful in
detecting outliers. The first use of the smallest PC in the
literature was done by Gnanadesikan and Wilk 1969 \cite{gnanadesikan1969data}.
Based on this work, Gnanadesikan \cite{gnanadesikan1972robust} stated that ``with
p-dimensional data, the projection onto the smallest principal
component would be relevant for studying the deviation of an
observation from a hyperplane of closest fit''. More recently,
Izenman and Shen used the smallest kernel principal components
for outlier detection as a generalization of the linear case \cite{izenman2009outlier}. Alakkari et al. found that the least significant eigenface can be used as a basis for discriminating between face and non-face images \cite{alakkari2015investigation}. In Partial Differential Equations, many systems are solved by seeking a hyperplane that is constituted of the entire solution. This is known as the method of characteristics.

\section{Concepts}

	The standard approach to PCA is as follows. Given data samples $X=[x_{1}\, x_{2}\cdots x_{n}]\in\mathbb{R}^{d\times n}$,
	where each sample is in column vector format, the covariance matrix
	is defined as
	\begin{equation}
	C=\frac{1}{n-1}\sum_{i=1}^{n}\left(x_{i}-\bar{x}\right)\left(x_{i}-\bar{x}\right)^{T},
	\label{eq:covariance}
	\end{equation}
	where $\bar{x}$ is the sample mean. In the sequel of this paper, we will assume that all samples are centered and hence there is no need to subtract the sample mean explicitly. After computing the covariance matrix, we can find the optimal low-dimensional bases that cover most variability in samples by extracting the significant eigenvectors of the
	covariance matrix $C$. Eigenvectors are extracted by solving the following eigenvalue
	equation
	\begin{equation}
	\left(C-\lambda I\right)v=0;\, v^{T}v=1,\label{eq:characteristiceqn}
	\end{equation}
	where $v\in\mathbb{R}^{d}$ is the eigenvector and $\lambda$ is its
	corresponding eigenvalue. Eigenvalues describe the variance maintained by the corresponding
	eigenvectors. Hence, we are interested in the subset of eigenvectors that have
	the highest eigenvalues $V=[v_{1}\, v_{2}\cdots v_{p}];\, p\ll n$. Then we encode a
	given sample $x$ using its $p$-dimensional projection values (referred to as \emph{scores}) as
	follows
	\begin{equation}\label{eq:projection}
	W=V^{T}x.
	\end{equation}
	We can then reconstruct the sample as follows
	\begin{equation}\label{eq:reconstruction}
	x_{reconstructed}=VW.
	\end{equation}
	One advantage of PCA is the low computational complexity when it comes to encoding and reconstructing samples.
	
	\subsection*{Duality in PCA}
	Since in the case of $n\ll d$, $C$ will be of rank $n-1$ and hence
	there are only $n-1$ eigenvectors that can be extracted from Eq. \eqref{eq:characteristiceqn}
	and since $C$ is of size $d\times d$, solving Eq. \eqref{eq:characteristiceqn}
	becomes computationally expensive. We can find such eigenvectors from the dual eigenspace by computing the $n\times n$ matrix $X^{T}X$ and then
	solving the eigenvalue problem
	\begin{equation}
	\left(X^{T}X-(n-1)\lambda I\right)v_{dual}=0
	\end{equation}
	\begin{equation}
	\Rightarrow X^{T}Xv_{dual}=(n-1)\lambda v_{dual};\, v_{dual}^{T}v_{dual}=1.\label{eq:eigenvalue2}
	\end{equation}

	Here, for simplicity, we assumed that the sample mean of $X$ is the zero vector. After extracting the dual eigenvectors, one can note that by multiplying
	each side of Eq. \eqref{eq:eigenvalue2} by $X$, we have
	\[
	XX^{T}Xv_{dual}=(n-1)\lambda Xv_{dual}
	\]
	\[
	\Rightarrow\frac{1}{n-1}XX^{T}\left(Xv_{dual}\right)=\lambda\left(Xv_{dual}\right)
	\]
	\[
	\Rightarrow C\left(Xv_{dual}\right)=\lambda\left(Xv_{dual}\right)
	\]
	\[
	\Rightarrow\left(C-\lambda I\right)\left(Xv_{dual}\right)=0
	\]

	which implies that 
	\begin{equation}
	v=Xv_{dual}.\label{eq:vdual}
	\end{equation}

	Thus, when $n\ll d$, we only need to extract the dual eigenvectors
	using Eq. \eqref{eq:eigenvalue2} and then compute the real eigenvectors
	using Eq. \eqref{eq:vdual}. Only the first few eigenvectors $V_{p}=[v_{1}\: v_{2}\ldots v_{p}],\: p\ll n\ll d$
	will be chosen to represent the eigenspace, those with larger eigenvalues.

\section{Adaptive PCA Algorithm}

The main premise of our algorithm is based on the fact that an eigenvector is actually a weighted sum of the input samples. 
We can show that by rewriting Eq. \eqref{eq:characteristiceqn} as follows \[
\left(\frac{1}{n-1}\sum_{i=1}^{n}x_{i}x_{i}^{T}-\lambda I\right)v=0
\]
\[
\Rightarrow\frac{1}{n-1}\sum_{i=1}^{n}x_{i}x_{i}^{T}v-\lambda v=0
\]
\[
\Rightarrow\frac{1}{n-1}\sum_{i=1}^{n}x_{i}\left\langle x_{i},v\right\rangle -\lambda v=0
\]
\[
\Rightarrow v=\frac{1}{\lambda\left(n-1\right)}\sum_{i=1}^{n}\left\langle x_{i},v\right\rangle x_{i}.
\]
A first guess for an update formula given new time-step $x_{n+1}$ would be \[
v^{t+1}=v^{t}+\left\langle x_{n+1},v^{t}\right\rangle x_{n+1};\;v^{t+1}=\frac{v^{t+1}}{\left\Vert v^{t+1}\right\Vert }.
\] This is similar to Oja's update scheme mentioned in the background section. The problem with this formula is that it assumes the weightings of previous samples are fixed. As the eigenvector is updated for each new time-step, the weights of previous samples should also be adjusted according to their projections on the updated eigenvector. The change in weights will be proportional to the correlations between previous samples and the new time-step. In our algorithm we used the following update rule \begin{align*} 
v^{t+1}&=v^{t}+\left(\sum_{j=1}^{n}\left\langle v^{t},x_{j}\right\rangle \left\langle x_{j},x_{n+1}\right\rangle ^{2}x_{j}\right)+\left\langle v^{t},x_{n+1}\right\rangle \left(\sum_{j=1}^{n+1}\left\langle x_{j},x_{n+1}\right\rangle \right)^{2}x_{n+1}\\
&=v^{t}+\left(\sum_{j=1}^{n}\left\langle v^{t},x_{j}\right\rangle \left\langle x_{j},x_{n+1}\right\rangle ^{2}x_{j}\right)\\
&\;\;\;\;\;\;\;\;+\left\langle v^{t},x_{n+1}\right\rangle\left(\sum_{j=1}^{n+1}\sum_{i=1}^{n+1}\left\langle x_{j},x_{n+1}\right\rangle \left\langle x_{i},x_{n+1}\right\rangle \right)x_{n+1};\;v^{t+1}=\frac{v^{t+1}}{\left\Vert v^{t+1}\right\Vert }.
\end{align*}  Unlike Oja's method, this is an online scheme that adapts weightings of all previous samples based on the squared dot product with the new time-step. In addition, the new time-step is weighted based on the sum of all second order dot products $\left\{ \left\langle x_{i},x_{n+1}\right\rangle .\left\langle x_{j},x_{n+1}\right\rangle \right\} _{i,j=1}^{n+1}$ multiplied by new time-step's score~$\left\langle v^{t},x_{n+1}\right\rangle$. Since for each eigenvector, we are computing the correlations (dot products) between the new time-step $x_{n+1}$ and all $n$ previous samples and considering that scores (weights) of previous samples $\left\{ \left\langle v^{t},x_{j}\right\rangle \right\} _{j=1}^{n}$ are computed in the previous iteration, this requires a time complexity of  $\mathcal{O}\left(n\right)$ dot products per eigenvector per new time-step. 

The full pseudo-code of our algorithm is shown in Algorithm \ref{Adaptive_PCA}. There are two parameters used in our algorithm: $space\_limit$ which specifies the maximal number of significant eigenvectors to compute and $processing\_limit$ which specifies the maximal number of dot products to compute per new time-step per eigenvector. As we mentioned earlier, our algorithm is capable of finding all eigenvectors of the data. In order to compute the full dimensional eigenspace deterministically, we set $space\_limit=\min(d,n)$ and $processing\_limit\gg n$ where $d$ is the total number of dimensions per sample and $n$ is the current number of samples. In its full-dimensional mode, our algorithm starts with two time-steps with $\frac{x_{2}-x_{1}}{\left\Vert x_{2}-x_{1}\right\Vert }$ as the initial eigenvector and ends with the full-dimensional eigenspace of the data. Line \ref{update_rule} of the algorithm includes the general update rule. Line \ref{share_experience} is used particularly for the limited processing mode (stochastic mode) to stress the shared information learned by $v^{t}$ and $v^{t+1}$. Line \ref{orthogonalize} performs a Gram-Schmidt process to ensure that following update terms will be orthogonal to updated eigenvector. After finishing the loop, $\tilde{X}$ will constitute the $n$th eigenvector since it will be perpendicular to all $n-1$ updated components.

\begin{algorithm}[t]
    \SetKwInOut{Input}{Input}
    \SetKwInOut{Output}{Output}
    
    \For{each new time-step $x_{n+1}$}
    {
        $X=\left[X,\;x_{n+1}\right]$\;
        $\tilde{X}=X$\;
        \eIf{$n>processing\_limit$}
        {
            $indices=\text{rand}\left(n,processing\_limit\right)$\; 
        }
        {
            $indices=1:n$\;  
        }
        
        \For{$i=1:\left(\min\left(n,space\_limit\right)-1\right)$}
        {
           $\tilde{v}=v_{i}+\left(\sum_{j=indices}\left\langle v_{i},\tilde{x}_{j}\right\rangle \left\langle \tilde{x}_{j},\tilde{x}_{n+1}\right\rangle ^{2}\tilde{x}_{j}\right)+\left\langle v_{i},\tilde{x}_{n+1}\right\rangle \left(\left(\sum_{j=indices}\left\langle \tilde{x}_{j},\tilde{x}_{n+1}\right\rangle \right)+\left\langle \tilde{x}_{n+1},\tilde{x}_{n+1}\right\rangle \right)^{2}\tilde{x}_{n+1}$\; \label{update_rule}
           $v_{i}=\tilde{v}+\left\langle \tilde{v},v_{i}\right\rangle v_{i}$\; \label{share_experience}
           $v_{i}=\frac{v_{i}}{\left\Vert v_{i}\right\Vert }$\;
           $\tilde{X}_{indices\cup\left\{ n+1\right\} }=\tilde{X}_{indices\cup\left\{ n+1\right\} }-v_{i}\left(v_{i}^{T}\tilde{X}_{indices\cup\left\{ n+1\right\} }\right)$\; \label{orthogonalize}
        }
        $v_{\min\left(n,space\_limit\right)}=\sum_{j=indices\cup\left\{ n+1\right\}}\tilde{x}_{j}$\;
        $n=n+1$\;
    }    
    \caption{Adaptive PCA}
    \label{Adaptive_PCA}
\end{algorithm}

\subsection{Limited-Dimensional Adaptive PCA}

In the limited dimensional mode of our algorithm, we set a maximal number of eigenvectors to update/compute per new time-step using the \emph{space\_limit} parameter. Since this parameter value is constant throughout the execution of our algorithm, this will bound the time complexity to $\mathcal{O}\left(space\_limit \times n\right)=\mathcal{O}\left(n\right)$ dot products per new time-step.

\subsection{Limited-Dimensional Adaptive PCA in Stochastic Mode}

In the stochastic mode, we specify a maximal number of dot products to be computed per new time-step per eigenvector. This happens when $n>processing\_limit$. In this case, we choose $processing\_limit$ uniformly distributed random samples to compute their dot products with the new time-step. This will further bound the time complexity to $\mathcal{O}\left(space\_limit \times processing\_limit\right)=\mathcal{O}\left(1\right)$ dot products per new time-step. Considering that our algorithm does not require the computation of the covariance matrix, the full time complexity of PCA in the stochastic mode will be $\mathcal{O}\left(n\right)$ dot products (after processing all time-steps of the input data).

\section{Experimental Results}
We applied our algorithm on six time-varying datasets of different physical phenomena. The first dataset studies the stages of a supernova during a period of less than one second after a star's core collapses~\cite{Supernova}. The second  dataset studies the fluid dynamics in turbulent vortex in a 3D area of the fluid~\cite{TurbulentVortex}. The third dataset shows the evolution of a splash caused after a drop impacts on a liquid surface~\cite{thoraval2012karman}. The fourth dataset was generated to analyze the chaotic path of a drop of silicone oil bouncing on the surface of a vibrating bath of the same material~\cite{perrard2016wave}. The fifth experimental data shows the unusual behaviour of some particles self-organized into spirals after rotational fluid flow~\cite{pushkin2011ordering}. Finally, the sixth dataset shows the behaviour of nitrogen bubbles cascading down the side of a glass of Guinness (dark beer), which has been well-investigated in a number of papers~\cite{benilov2013bubbles,power2009initiation}. Table~\ref{tab:summary} summarizes the properties of each dataset. The first two datasets are in the form a 3D scalar field (i.e. voxels datasets). The remaining four datasets are in video format and were adapted from original sources by converting to greyscale video frames and cropping these frames to a segment of interest (the most highly varying part of the video sequence). 
The adapted datasets can be obtained by emailing the authors.

\begin{table}[t]
\caption{\label{tab:summary}
            Summary of the datasets used in our experiments.}

\centering{}%
\begin{tabular}{|c|c|c|c|}
\hline 
\textbf{\scriptsize{}dataset/experiment name} & \textbf{\scriptsize{}data type} & \textbf{\scriptsize{}time-step resolution} & \textbf{\scriptsize{}number of time-steps}\tabularnewline
\hline 
\hline 
\textbf{\scriptsize{}Supernova} & {\scriptsize{}3D volumes} & {\scriptsize{}$432^{3}$ voxels} & {\scriptsize{}60}\tabularnewline
\hline 
\textbf{\scriptsize{}Turbulent Vortex} & {\scriptsize{}3D volumes} & {\scriptsize{}$128^{3}$ voxels} & {\scriptsize{}100}\tabularnewline
\hline 
\textbf{\scriptsize{}Droplet Impact on Liquid Surface} & {\scriptsize{}grayscale video} & {\scriptsize{}$240\times312$ pixels} & {\scriptsize{}100}\tabularnewline
\hline 
\textbf{\scriptsize{}Bouncing Silicone Drop} & {\scriptsize{}grayscale video} & {\scriptsize{}$300\times640$ pixels} & {\scriptsize{}300}\tabularnewline
\hline 
\textbf{\scriptsize{}Self Organized Particles} & {\scriptsize{}grayscale video} & {\scriptsize{}$54\times152$ pixels} & {\scriptsize{}650}\tabularnewline
\hline 
\textbf{\scriptsize{}Guinness Cascade} & {\scriptsize{}grayscale video} & {\scriptsize{}$271\times131$ pixels} & {\scriptsize{}1,100}\tabularnewline
\hline 
\end{tabular}
\end{table}

We compare the performance of our algorithm with standard PCA in terms of explained variance curves. The standard PCA results are generated using the \emph{pcacov} function in MATLAB~\cite{MATLAB:2017}. Since we are dealing with cases where $n\ll d$, it is typical to use the dual covariance matrix $X^{T}X$ for standard PCA. For the adaptive PCA, we incrementally update the eigenvectors until reaching the last time-step as in Algorithm~\ref{Adaptive_PCA}.

\begin{figure}[t]

			\begin{minipage}{0.33\textwidth}
			\centering
			\includegraphics[width=1.0\linewidth]{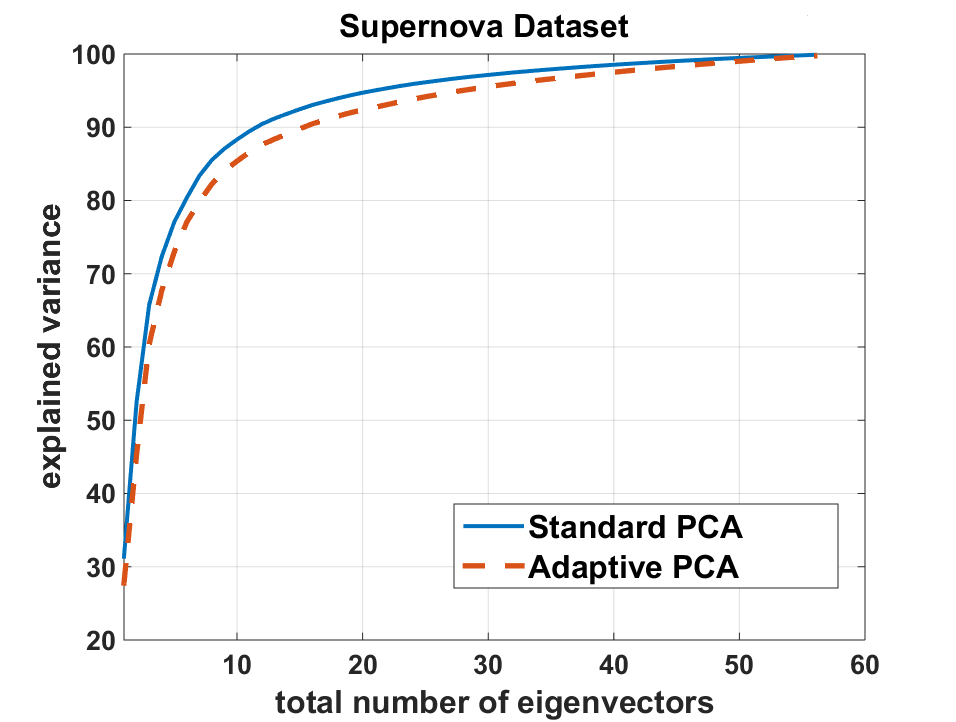}
			
		\end{minipage}
		\begin{minipage}{0.33\textwidth}
			\centering
			\includegraphics[width=1.0\linewidth]{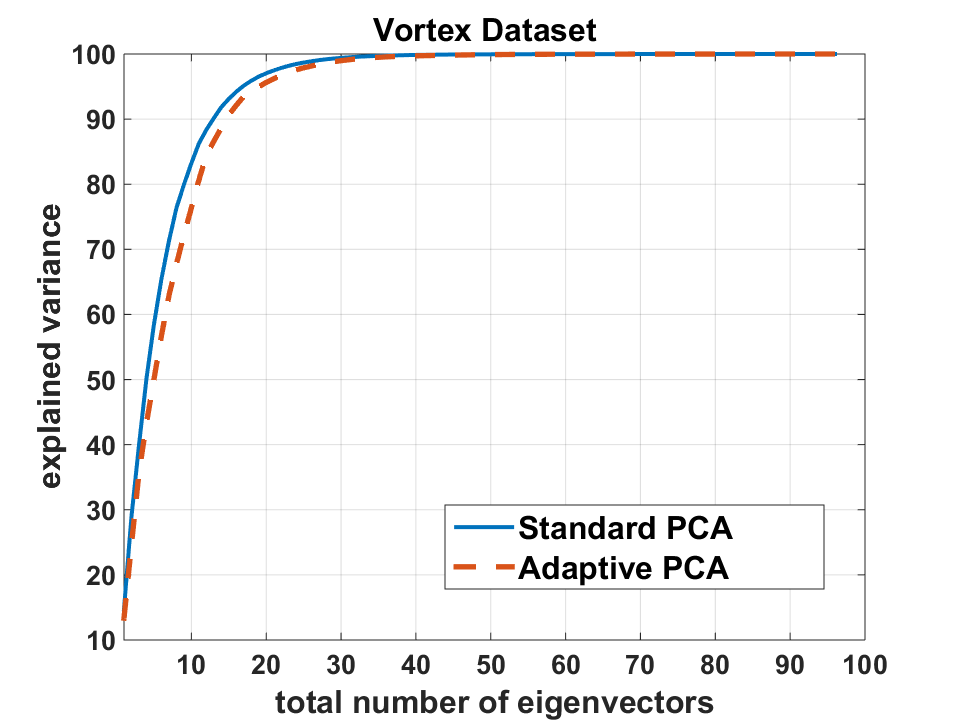}
		\end{minipage}
		\begin{minipage}{0.328\textwidth}
			\centering
			\includegraphics[width=1.0\linewidth]{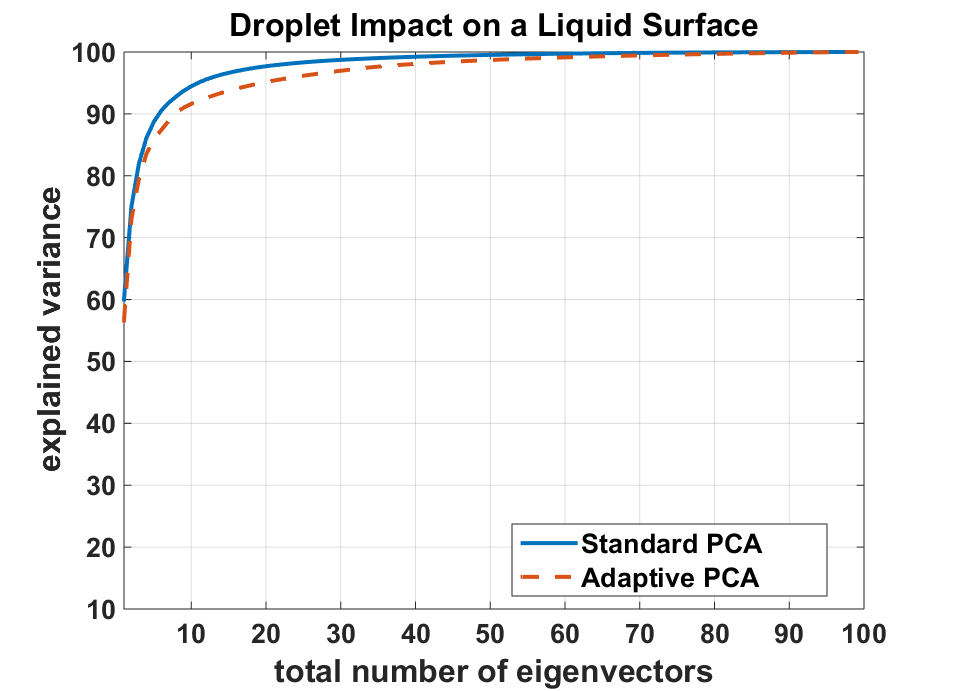}
		\end{minipage}\\
		\begin{minipage}{0.33\textwidth}
			\centering
			\includegraphics[width=1.0\linewidth]{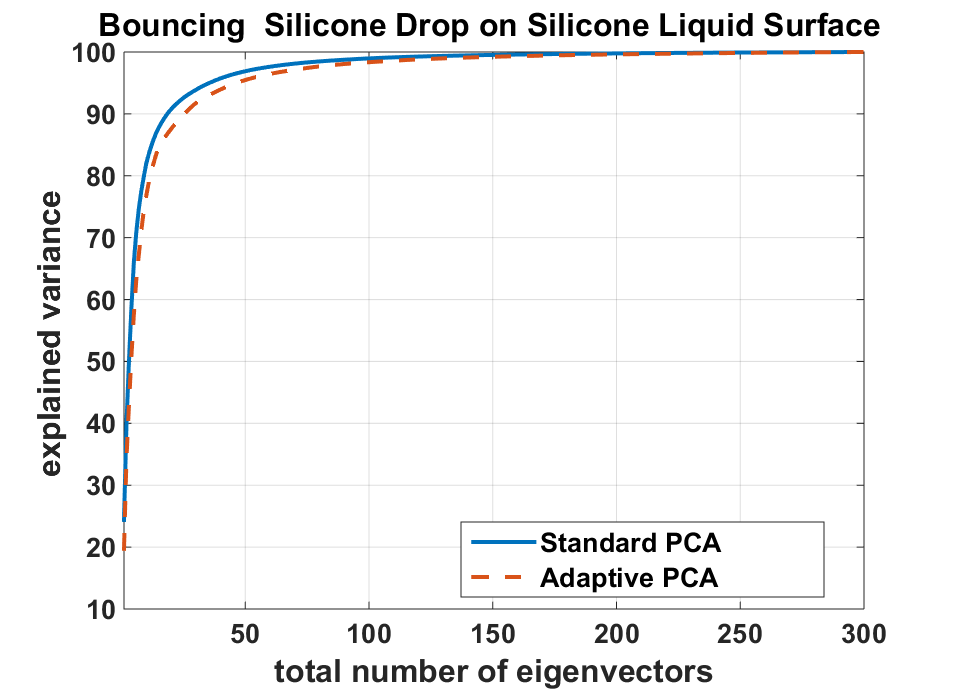}
		\end{minipage}
		\begin{minipage}{0.33\textwidth}
			\centering
			\includegraphics[width=1.0\linewidth]{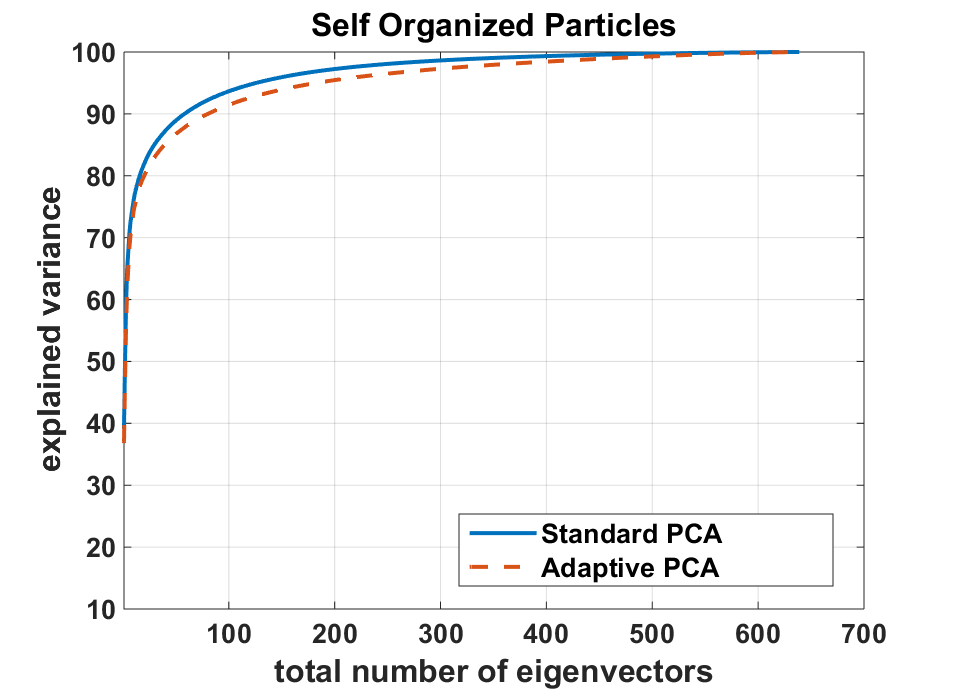}
		\end{minipage}
		\begin{minipage}{0.328\textwidth}
			\centering
			\includegraphics[width=1.0\linewidth]{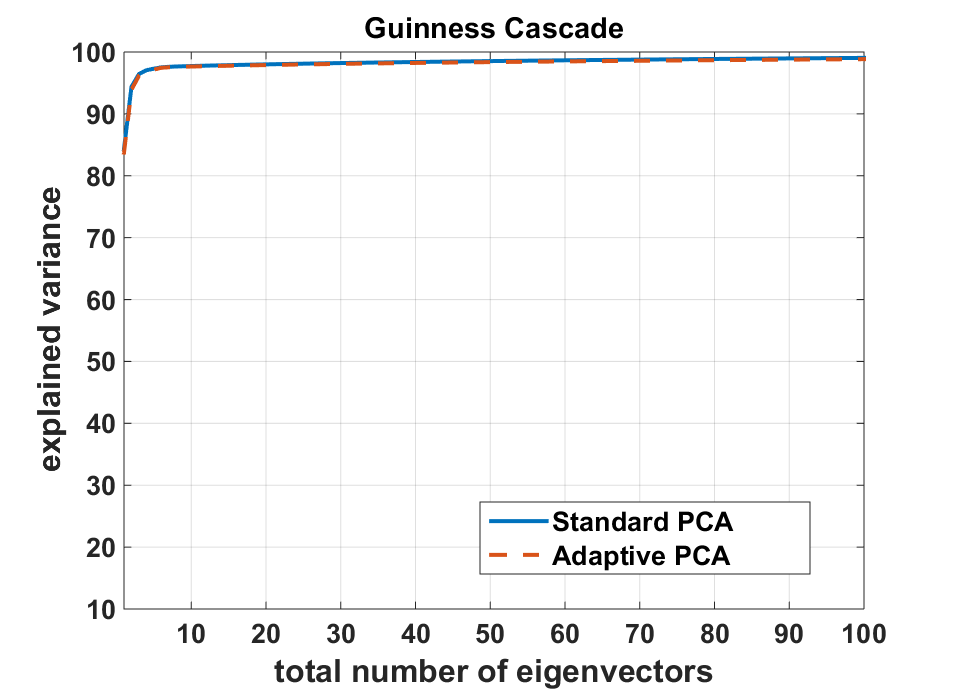}
		\end{minipage}
		
			\caption{\label{fig:full-mode}
				Explained variance curves of standard PCA and full-dimesional adaptive PCA for each dataset.}
		\end{figure} 
		
We first do a comparison between the full-dimensional eigenspace computed using each technique. This is to stress the capability of our approach of finding all eigenvectors of the given datasets. Figure~\ref{fig:full-mode} shows the explained variance curves for each dataset. It is very clear that our algorithm provides an excellent approximation to the original full-dimensional eigenspace. For all datasets, the gap between the two curves does not exceed $2\%$. It is also interesting to note how well both techniques were able to learn the guinness cascade phenomenon, where $98\%$ of the variability was covered by only the first 20 eigenvectors.

\begin{figure}[t]

			\begin{minipage}{0.33\textwidth}
			\centering
			\includegraphics[width=1.0\linewidth]{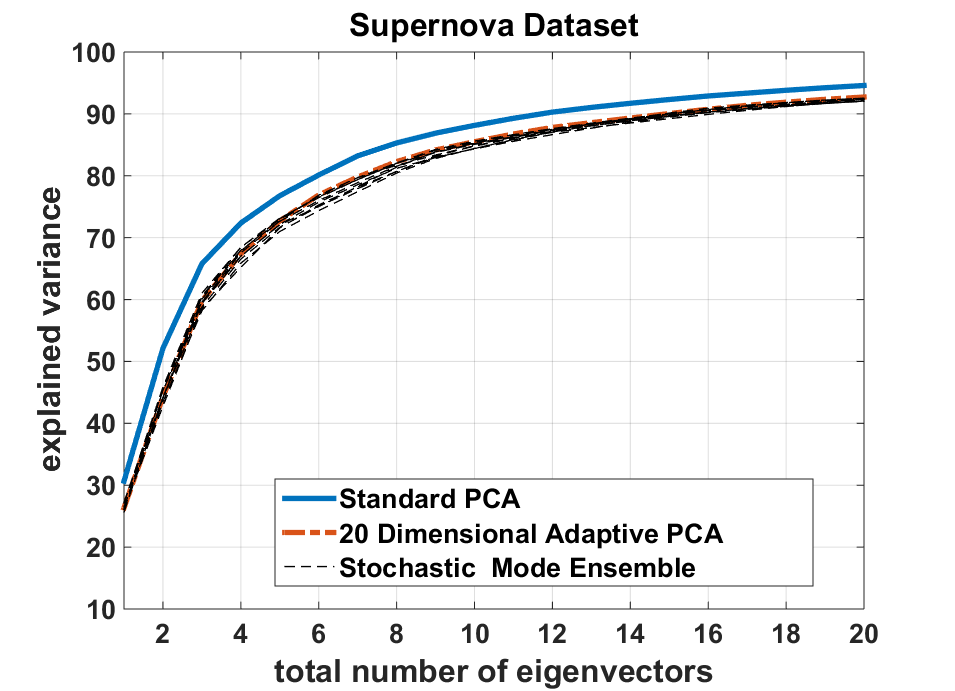}
			
		\end{minipage}%
		\begin{minipage}{0.33\textwidth}
			\centering
			\includegraphics[width=1.0\linewidth]{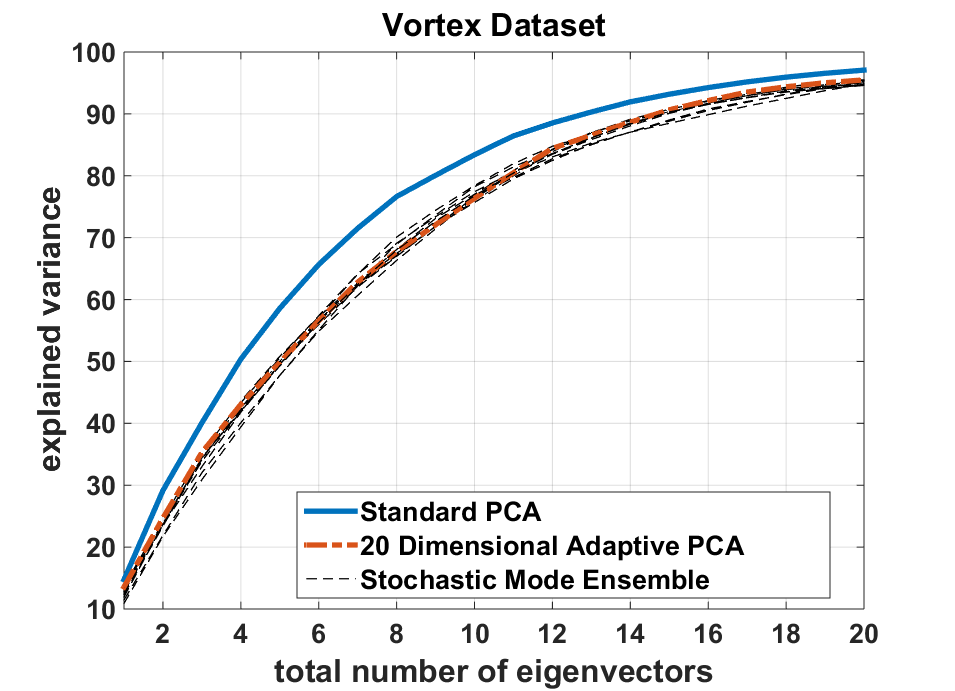}
		\end{minipage}
		\begin{minipage}{0.328\textwidth}
			\centering
			\includegraphics[width=1.0\linewidth]{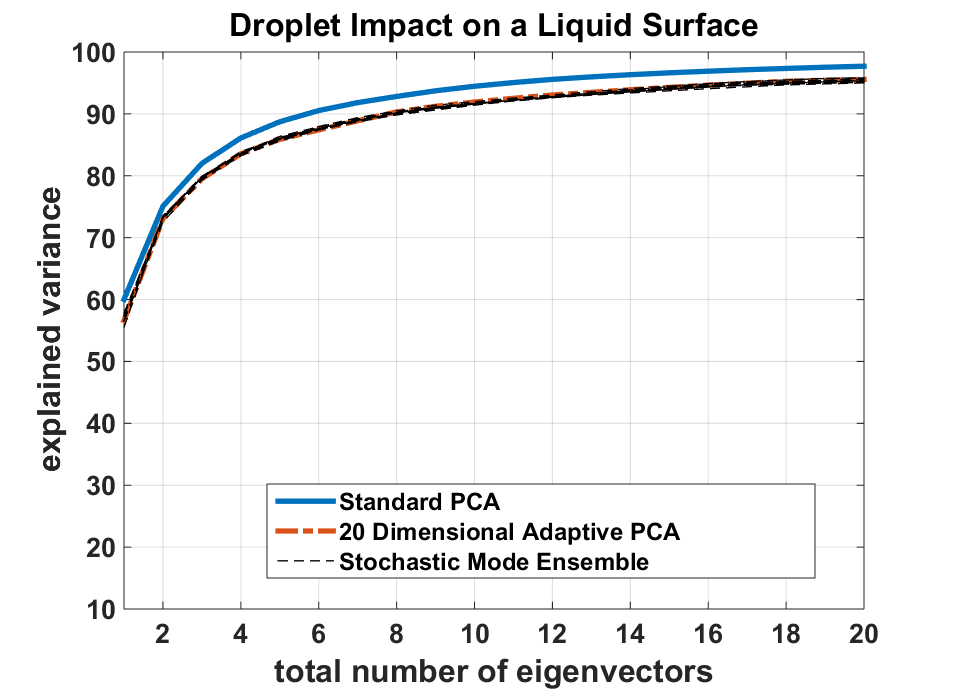}
		\end{minipage}
		\begin{minipage}{0.33\textwidth}
			\centering
			\includegraphics[width=1.0\linewidth]{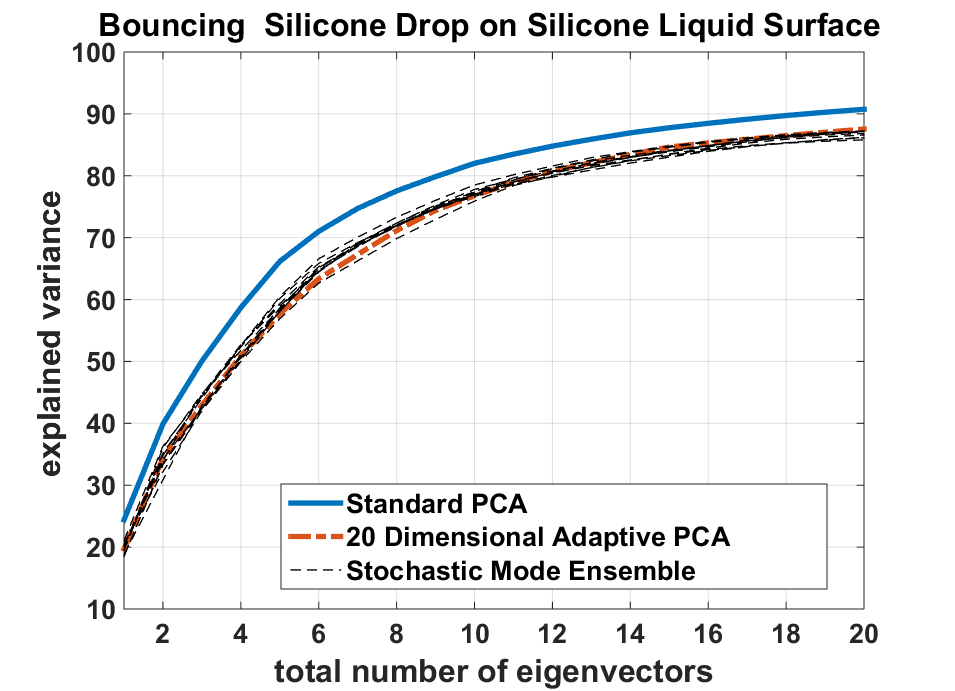}
		\end{minipage}
		\begin{minipage}{0.33\textwidth}
			\centering
			\includegraphics[width=1.0\linewidth]{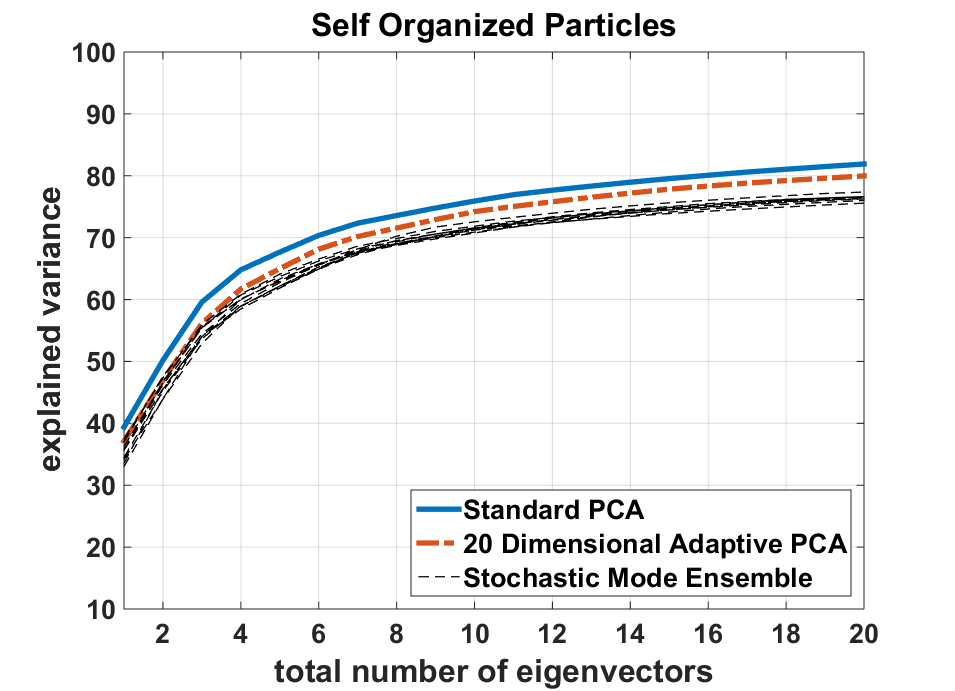}
		\end{minipage}
		\begin{minipage}{0.328\textwidth}
			\centering
			\includegraphics[width=1.0\linewidth]{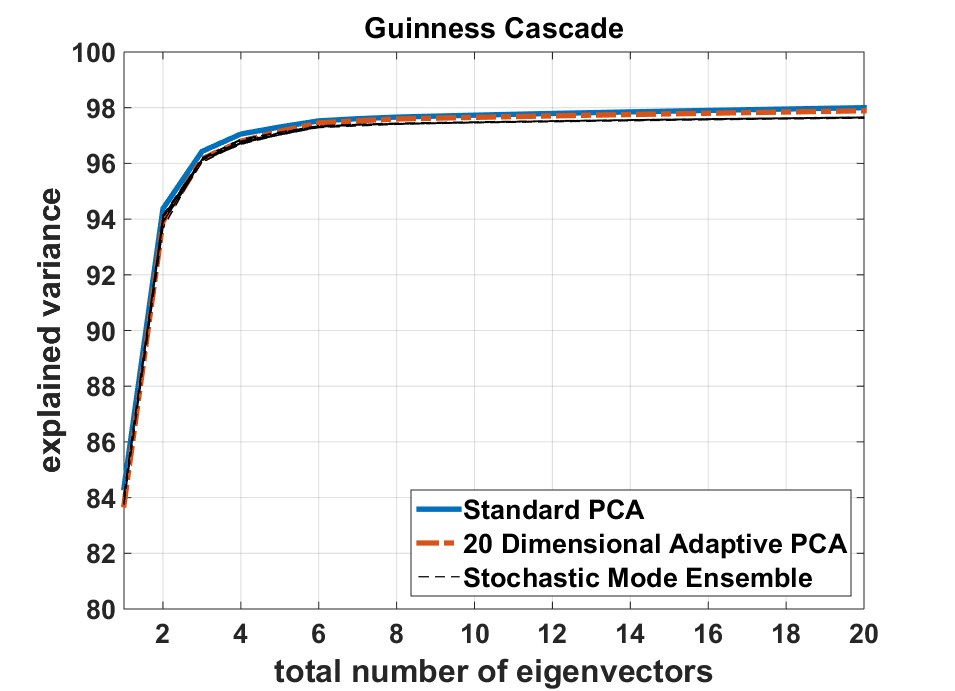}
		\end{minipage}
		
			\caption{\label{fig:20D-mode}
				Explained variance curves of standard PCA and 20-dimensional adaptive PCA in both deterministic and stochastic modes for each dataset.}
		\end{figure}

Next we compute the limited-dimensional eigenspace for each dataset mainly to compare performance between deterministic and stochastic modes of our algorithm. Figure~\ref{fig:20D-mode} shows the performance of the 20-dimensional adaptive PCA. To test the stochastic mode performance, we applied 10 runs of our algorithm where in each run we set $processing\_ limit$ to 40. With much lower number of computations, the stochastic runs  achieve almost the same performance as the deterministic mode. The only difference one can note is in the \emph{Self Organized Particles} experiment where the stochastic runs provide mean explained variance of $77\%$ while the deterministic mode covers $80\%$ of variability.
		
\section{Conclusion}
\begin{figure}[h]
		\begin{minipage}{.5\textwidth}
			\centering
			\includegraphics[width=1.0\linewidth]{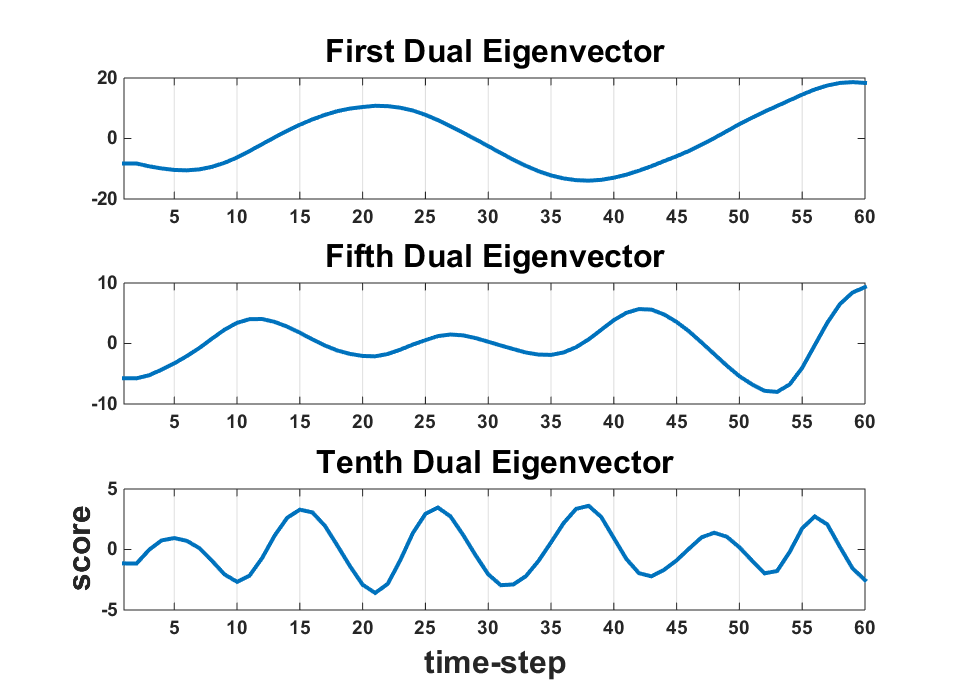}
			
		\end{minipage}%
		\begin{minipage}{.5\textwidth}
			\centering
			\includegraphics[width=1.0\linewidth]{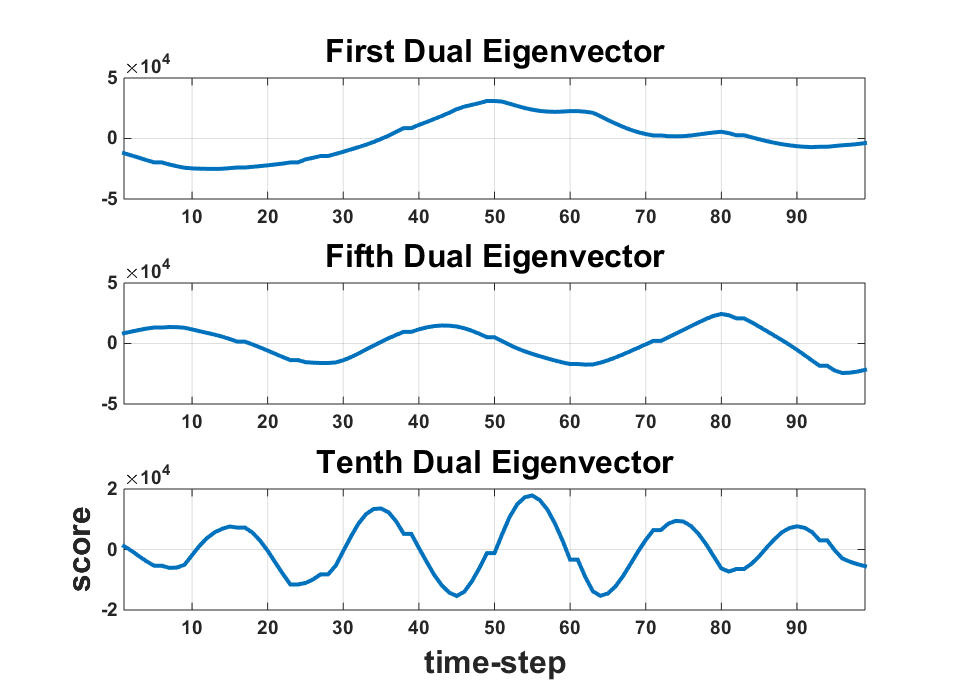}
		\end{minipage}
		
			\caption{\label{fig:dual_eigenspace}
				Three time-dependent eigenvectors for Supernova (left) and Vortex (right) datasets.}
		\end{figure}

In this paper, we presented a deterministic scheme that finds the eigenspace of sequential data incrementally with linear time complexity growth. Our model is a generalization of Oja's method with the following two main advantages. In our approach, the eigenvectors are updated in an online manner (one-step update per eigenvector) unlike Oja's method which is applied in an iterative manner per eigenvector. Secondly, our model considers all previous samples in its update formula whereas Oja's method considers only the most recent time-step in its update rule. Since our algorithm considers all second order correlations between samples, this provides an intensive learning scheme that better resembles the quadratic nature of standard PCA. 
In the limited-computations mode of our algorithm (stochastic mode), the eigenvectors are adapted according to the pattern learned from limited population ensembles. Our experiments have shown that the stochastic mode provides the same performance as the deterministic mode with much lower number of computations. 
Our technique serves as a robust modeling tool for complex time-dependent systems that decomposes the systems temporal behaviour using orthogonal time-dependent functions which correspond to the dual eigenspace. This can be expressed as follows \[
\vec{S}_{t}=\sum_{i=1}^{p}v_{i}f_{i}\left(t\right)=\sum_{i=1}^{p}v_{i}\left(v_{i}^{T}x_{t}\right).
\] Figures~\ref{fig:dual_eigenspace} shows the time-dependent functions of the first, fifth and tenth eigenvectors for the Supernova and Vortex datasets. One can note that the higher significance eigenfunctions have lower frequency with higher amplitude.  By interpolating these functions, we can analyze the system behavior in continuous time. In terms of future work, it would be interesting to know the performance of our algorithm using different distributions of previous samples in the stochastic mode. 
In many systems, the recent samples have higher priority than older ones, such as in CCTV surveillance applications where the records are saved for a limited period of time.

\subsubsection*{Acknowledgments}

This research has been conducted with the financial support of Science Foundation Ireland (SFI) under Grant Number 13/IA/1895.


\bibliographystyle{abbrv}

\bibliography{template}

\end{document}